\documentclass[preprint,12pt]{elsarticle}

\usepackage{amssymb}
\usepackage{amsmath}

\usepackage{graphicx}
\usepackage[labelsep=period]{caption} 
\usepackage{subcaption}

\usepackage{booktabs}
\usepackage{multirow}
\usepackage{array}
\usepackage{threeparttable}
\usepackage{float}
\usepackage{makecell}
\usepackage{tabularx}

\usepackage{footmisc} 

\usepackage{hyperref}
\usepackage{url}

\journal{ISPRS Journal of Photogrammetry and Remote Sensing}

\begin{document}

\begin{frontmatter}

\title{SpecSwin3D: Generating Hyperspectral Imagery from Multispectral Data via Transformer Networks}

\author[1]{Tang Sui}
\author[1]{Songxi Yang}
\author[1]{Qunying Huang\corref{cor1}}

\cortext[cor1]{Corresponding author: qhuang46@wisc.edu}

\affiliation[1]{organization={Department of Geography, University of Wisconsin-Madison},
                addressline={Science Hall}, 
                city={Madison},
                postcode={53706}, 
                state={Wisconsin},
                country={USA}}

\begin{abstract}
Multispectral and hyperspectral imagery are widely used in agriculture, environmental monitoring, and urban planning due to their complementary spatial and spectral characteristics. A fundamental trade-off persists: multispectral imagery offers high spatial but limited spectral resolution, while hyperspectral imagery provides rich spectra at lower spatial resolution. Prior hyperspectral generation approaches (e.g., pan-sharpening variants, matrix factorization, CNNs) often struggle to jointly preserve spatial detail and spectral fidelity. In response, we propose \textbf{SpecSwin3D}, a transformer-based model that generates hyperspectral imagery from multispectral inputs while preserving both spatial and spectral quality. Specifically, SpecSwin3D takes five multispectral bands as input and reconstructs 224 hyperspectral bands at the same spatial resolution. In addition, we observe that reconstruction errors grow for hyperspectral bands spectrally distant from the input bands. To address this, we introduce a \textbf{cascade training strategy} that progressively expands the spectral range to stabilize learning and improve fidelity. Moreover, we design an optimized band sequence that strategically repeats and orders the five selected multispectral bands to better capture pairwise relations within a 3D shifted-window transformer framework. Quantitatively, our model achieves a PSNR of \textbf{35.82 dB}, SAM of \textbf{$2.40^{\circ}$}, and SSIM of \textbf{0.96}, outperforming the baseline MHF-Net by \textbf{+5.6 dB} in PSNR and reducing ERGAS by more than half. Beyond reconstruction, we further demonstrate the practical value of \textbf{SpecSwin3D} on two downstream tasks, including land use classification and burnt area segmentation. 
\end{abstract}

\begin{graphicalabstract}
  \includegraphics[width=1\textwidth]{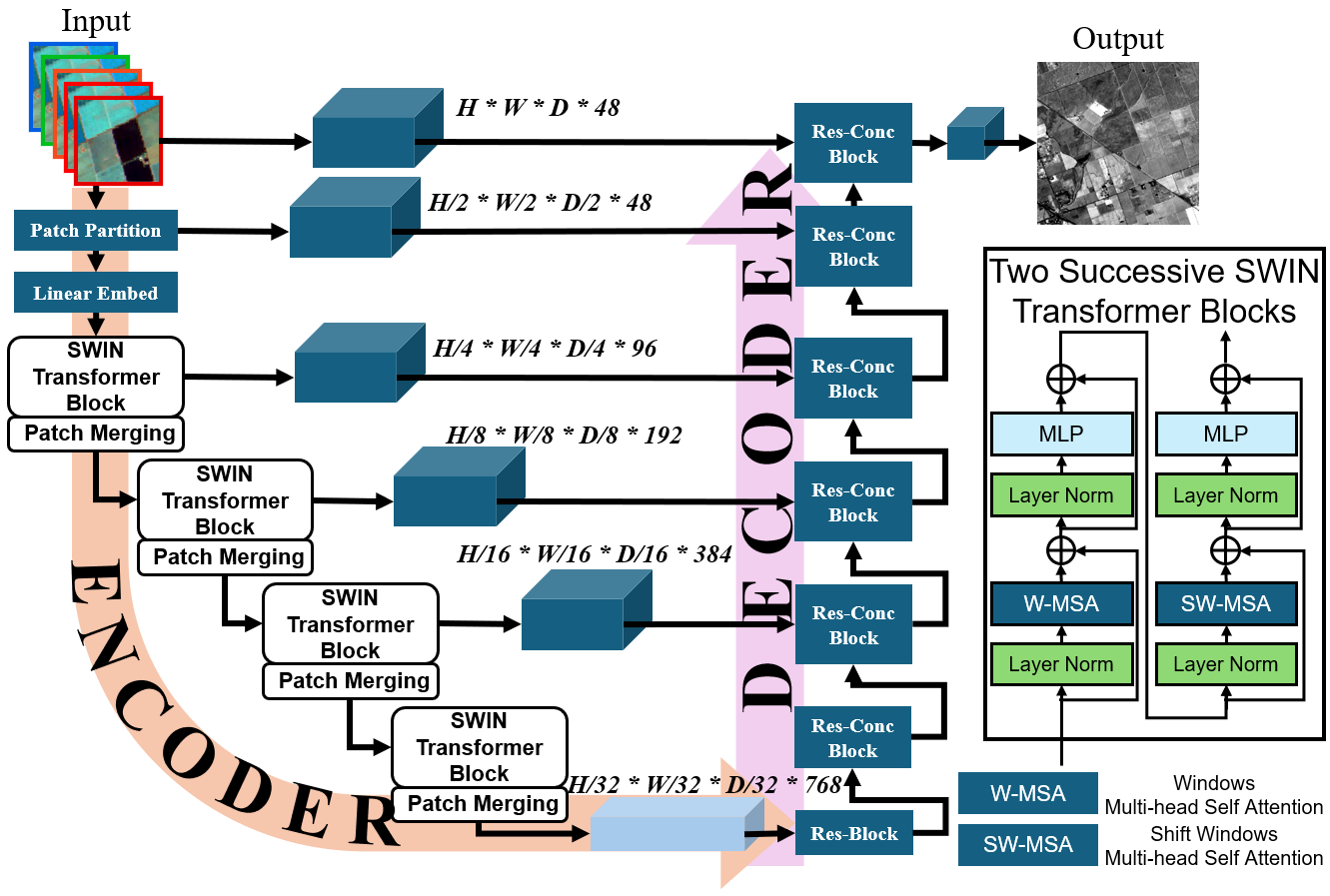}
\end{graphicalabstract}

\begin{highlights}
\item Propose SpecSwin3D, a 3D shifted-window Transformer for hyperspectral generation.
\item Introduce an optimized band sequence strategy to maximize inter-band interaction.
\item Present a cascade training strategy to balance performance and efficiency.
\item Validate and demonstrate the superior accuracy and robustness of SpecSwin3D using AVIRIS hyperspectral data.
\item Demonstrate the effectiveness of SpecSwin3D on downstream tasks using land use classification and burnt area segmentation.
\end{highlights}


\end{frontmatter}

\section{Introduction}
Multispectral and hyperspectral imagery are pivotal in diverse applications such as precision agriculture \cite{gevaert2015generation}, change detection \cite{kwan2019methods}, and land use classification \cite{vali2020deep}, owing to their ability to capture rich spatial and spectral information \cite{amigo2019hyperspectral}. Multispectral imagery typically consists of a few discrete spectral bands—such as blue, green, red, and near-infrared (400–900 nm)—and provides high spatial but limited spectral resolution. In contrast, hyperspectral imagery captures hundreds of narrow, contiguous bands across a broader spectral range (e.g., 400–2500 nm), offering finer spectral resolution at the cost of reduced spatial resolution \cite{yokoya2017hyperspectral}. This enhanced spectral granularity enables detailed discrimination of subtle material or biochemical differences, which is essential in applications such as monitoring plant traits in agriculture \cite{feng2021comprehensive,yang2017unmanned}, art conservation and archaeology \cite{liang2012advances}, food quality assessment \cite{gowen2007hyperspectral,feng2012application} and military surveillance \cite{briottet2006military,clark1995mapping}.
\par
Early methods for enhancing spectral or spatial resolution in remote sensing focused on pan-sharpening, which fuses high-resolution panchromatic imagery with lower-resolution multispectral data. While these approaches differ in objective and input data from modern hyperspectral imagery generation (HIG) techniques, they share a common foundation in statistical image fusion methods. Initial techniques often employed histogram matching to align intensity distributions between bands, aiming to minimize spectral distortion during the fusion process. Additionally, a number of transformation-based methods were proposed to improve fusion quality, including Intensity-Hue-Saturation \cite{carper1990use}, Principal Component Analysis \cite{kwarteng1989extracting}, and Gram-Schmidt orthogonalization \cite{laben2000process}. These approaches laid the groundwork for more advanced spectral-spatial fusion methods used in later HIG research.
\par
To extend the fusion process from multispectral and panchromatic imagery to hyperspectral and multispectral imagery, \cite{chen2014fusion} proposed a method that involved dividing the spectral bands of hyperspectral imagery into several groups based on their spectral coverage. Later, \cite{selva2015hyper} further advanced this approach by synthesizing high-resolution images for each spectral band of hyperspectral imagery through linear regression with high spatial resolution multispectral imagery.
\par
Matrix factorization (MF)–based approaches are mainly divided into sparse representation and low-rank methods. Sparse representation treats the spectral basis as an overcomplete dictionary, with methods like K-SVD and non-negative learning used to learn dictionaries, while coefficients are estimated through sparse coding algorithms \cite{dong2016hyperspectral}. Low-rank methods assume spectral signatures lie within a low-dimensional subspace and use techniques such as singular value decomposition to learn the spectral basis \cite{liu2020truncated}. Fusion methods can be categorized into three types: the first estimates spectral bases from low spatial resolution hyperspectral imagery and coefficients from high spatial resolution multispectral imagery, as seen in works like sparse MF \cite{huang2013spatial}; the second uses both hyperspectral imagery and multispectral imagery to estimate coefficients, employing regularization \cite{wei2015hyperspectral}; the third, such as coupled non-negative MF, alternates between updating spectral bases and coefficients \cite{lanaras2015hyperspectral}. Sparse representation methods model spectral redundancies well, while low-rank methods offer faster computation for large-scale fusion tasks \cite{yokoya2011coupled}.
\par
Convolutional neural networks (CNNs) have become popular in image processing for their efficiency and ability to learn image features from training data. \cite{dong2014learning} introduced the SRCNN model for single image super-resolution, which led to the development of CNN-based pan-sharpening methods that combine multispectral imagery with high spatial resolution panchromatic images, adaptable for HIG. These methods can be categorized into one-branch and two-branch approaches. One-branch methods, like PanNet \cite{yang2017pannet}, combine features from multispectral imagery and hyperspectral imagery and use a single CNN branch to predict high spatial resolution hyperspectral imagery. CNN-based methods often require sufficient training data, but the lack of it can limit performance. To address this, \cite{dian2020regularizing} introduced a CNN denoiser that handles various data types. Two-branch CNN-based approaches \cite{yang2018hyperspectral} use separate networks for spatial and spectral features, which are then combined.
\par
\begin{figure}[ht!]
\begin{center}
		\includegraphics[width=0.7\columnwidth]{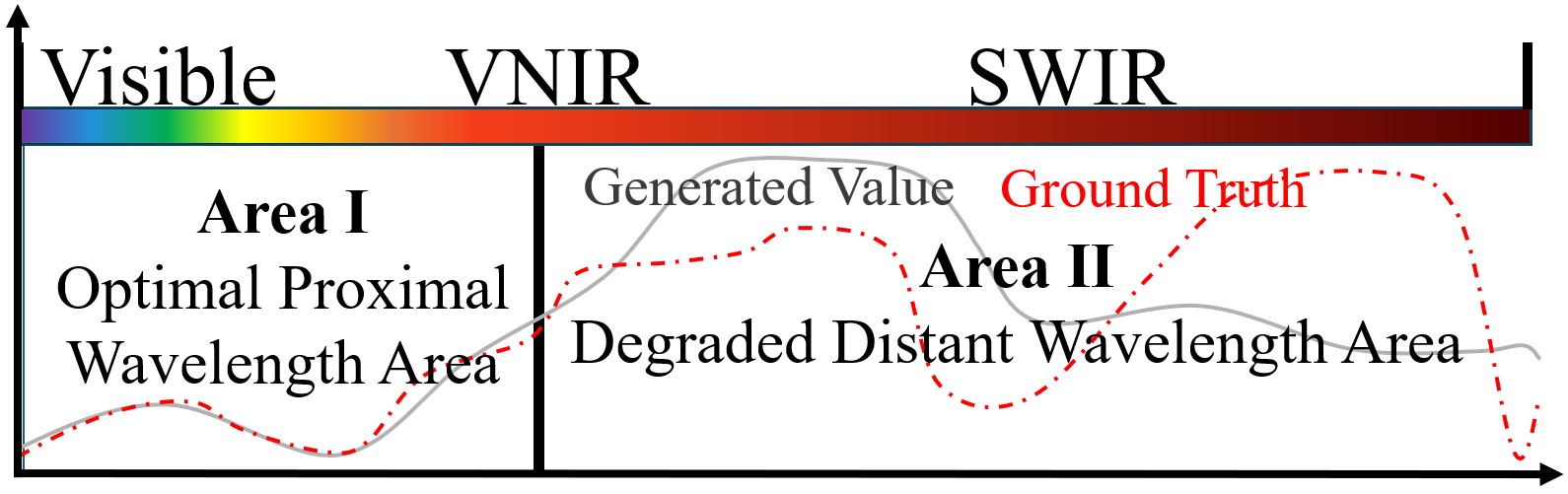}
	\caption{Preliminary spectral reconstruction analysis. 
The red curve represents the generated spectrum, while the gray curve denotes the ground truth. Two wavelength regions are highlighted: Area I, proximal to the input multispectral bands, where the generation values closely align with the ground truth, and Area II, more distant in wavelength, where the deviations become more pronounced.}
\label{fig:preliminary_result}
\end{center}
\end{figure}

However, these methods often struggle to simultaneously preserve both spatial and spectral quality. Pan-sharpening methods risk spectral distortion when enhancing spatial resolution during HIG \cite{dadrasjavan2018spectral}. MF methods involve costly optimization, are sensitive to difficult-to-configure parameters, and rely heavily on the observation model \cite{dian2021recent}. CNN-based methods, being data-driven, face generalization challenges. In addition, our preliminary work based on physics model \cite{xie2020mhf,wang2019deep,dong2021model} and transformer-based \cite{hu2022fusformer,jia2023multiscale} methods indicates that the reconstruction accuracy tends to decline for hyperspectral bands that are spectrally distant from the input multispectral bands. Shown in \autoref{fig:preliminary_result}, when the reconstructed bands are proximal to the input multispectral bands (i.e., Area I in \autoref{fig:preliminary_result}), their generated values closely align with the ground truth. However, when they are more distant in wavelength (i.e., Area II), the deviations become more pronounced.

\par
In response, this research proposes an innovative transformer-based deep learning model that generates hyperspectral bands from multispectral imagery while maintaining high spatial resolution with enhanced generalizability. The proposed model is developed using a comprehensive hyperspectral and multispectral dataset from California and Nevada. The dataset, sourced from the AVIRIS repository for model training, contains over 3,000 hyperspectral images (14 m spatial resolution, 224 bands: 400–2,500 nm).
\par
The contributions of this work include:
\begin{itemize}
  \setlength\itemsep{1em} 
  \setlength\parindent{2em} 
  \item An advanced transformer-based model, \textit{SpecSwin3D}, is proposed to capture complex spectral characteristics and fine spatial details from low spatial resolution hyperspectral imagery and high spatial resolution multispectral imagery, advancing HIG by combining the advantages of both imaging types.
  
  \item \textit{SpecSwin3D} is systematically compared against state-of-the-art HIG methods, and the experimental results indicate that it consistently outperforms these methods, demonstrating superior performance and robustness.

  \item An optimized band sequence strategy is proposed to maximize local inter-band interaction while ensuring full pairwise coverage in the spectral (depth) dimension, which enhances spectral feature extraction under a 3D shifted window framework.

  \item A cascade training strategy is introduced to progressively handle spectral bands with varying distances from the input multispectral bands, enabling more accurate hyperspectral band generation with reduced computational cost while preserving spectral fidelity and spatial detail.

  \item The practical value of \textit{SpecSwin3D} is validated on two downstream tasks—land use segmentation and burnt area detection—showing that the generated hyperspectral bands provide substantial benefits in both spatial and spectral domains.
\end{itemize}

\section{Background and Related Work}\label{Background_and_Related_Work}
Both hyperspectral imagery and multispectral imagery are crucial in fields such as medical imaging \cite{ortega2019use}, agriculture \cite{kamilaris2018deep}, ecosystem monitoring \cite{coppin2004review}, and change detection \cite{zhan2021tensor}. A major challenge arises from the design limitations of satellites, cameras, and other remote sensing technologies, which impose a trade-off between spatial and spectral resolution. As a result, modern satellite sensors often integrate hyperspectral imagery and multispectral imagery to capture a more comprehensive dataset for a given area. For example, Sentinel-2 provides multispectral imagery with 13 bands at up to 10 m spatial resolution, while Hyperion captures hyperspectral imagery with 220 bands (spanning 400–2500 nm) at a 30 m spatial resolution.
\par
Given the importance of both spatial and spectral information, finding effective methods to combine these two data types into a single image remains an ongoing and valuable research challenge. HIG refers to generating high spectral resolution data (i.e., hyperspectral imagery) from low spectral resolution inputs (i.e., multispectral imagery) by leveraging available spectral, spatial, and contextual information. This has become an important research topic in recent years. Existing approaches can be broadly categorized into two groups: statistical methods \cite{dian2019learning,dian2019hyperspectral,xu2019nonlocal} and deep learning methods \cite{xie2020mhf,wang2019deep,hu2022fusformer,dong2021model,jia2023multiscale}.
\subsection{Statistical Methods}\label{Statistical_Methods}
Traditional techniques for HIG can be categorized into two main approaches: pan-sharpening \cite{vivone2014critical,thomas2008synthesis} and MF–based methods \cite{simoes2014convex,wei2015hyperspectral,yokoya2011coupled,dian2021recent}.
\par
Pan-sharpening is an early technique for combining low spatial resolution multispectral imagery with high spatial resolution panchromatic images. \cite{vivone2014critical} summarized the most common pan-sharpening methods, such as component substitution. In this strategy, multispectral images are projected into a new domain—typically using a transformation that separates spatial structure from spectral features. The high-resolution spatial detail from the panchromatic image is then injected, and the inverse transformation is applied to reconstruct the image. The effectiveness of this approach improves with increasing correlation between the spectral bands of multispectral imagery and the spatial features in the panchromatic image \cite{thomas2008synthesis}.
\par
However, pan-sharpening methods are highly dependent on the spatial-spectral alignment between the source images. Any mismatch in spatial resolution, registration, or feature orientation can introduce artifacts, leading to spectral distortion and loss of structural fidelity in the fused imagery.
\par
A second class of approaches for HIG is based on MF. These methods aim to decompose the hyperspectral image into a spectral basis and corresponding coefficients through optimization. There are three typical formulations:
\begin{itemize}
  \setlength\itemsep{1em}
  \setlength\parindent{2em}

  \item \textbf{Full spectral-spatial split:} All spectral information is obtained from hyperspectral imagery, and all spatial information is extracted from multispectral imagery \cite{kawakami2011high,huang2013spatial,akhtar2014sparse}.

  \item \textbf{MAP-based coefficient estimation:} Some spatial information from hyperspectral imagery is used to estimate coefficients via maximum a posteriori estimation \cite{simoes2014convex,wei2015hyperspectral}.

  \item \textbf{Coupled dictionary learning:} Rather than using a fixed dictionary, the spectral basis and coefficients are alternately updated. A representative method in this category is coupled nonnegative matrix factorization \cite{yokoya2011coupled}.
\end{itemize}

While MF-based methods can enhance spatial resolution, they may introduce spectral distortion. For example, assuming that all spectral information comes exclusively from the hyperspectral image—an idealized assumption in the first approach—can neglect the spatial-spectral mixing present in real data \cite{dian2021recent}. The latter two methods involve complex optimization and may suffer from convergence issues \cite{wei2015hyperspectral}. Additionally, the third approach is highly sensitive to the spectral characteristics of observed materials, resulting in inconsistent textures or spectral artifacts across materials with diverse properties \cite{yokoya2011coupled}.

\subsection{Deep Learning Methods}\label{Deep Learning Methods}

As deep learning continues to advance, CNNs have already made significant contributions to HIG. CNN-based networks learn non-linear projection functions to model the correlation between hyperspectral imagery and multispectral imagery. Prior research has explored CNNs for spatial–spectral reconstruction \cite{zhang2020ssr}, cluster-based multi-branch architectures \cite{han2019hyperspectral}, and blind image fusion \cite{wang2019deep}. However, CNN-based methods typically perform well when generating hyperspectral bands that are close in wavelength to those in the multispectral imagery used for training, but their performance deteriorates for bands at significantly different wavelengths, as they often fail to capture inter-band relationships.
\par
\cite{zhang2020ssr} proposed a spatial–spectral reconstruction network based on an explainable CNN to fuse low spatial resolution hyperspectral imagery with high spatial resolution multispectral imagery. While this model incorporates spatial and spectral edge losses, its focus on local features and limited model capacity restricts the receptive field, making it difficult to capture broader contextual information.
\par
\cite{han2019hyperspectral} introduced a cluster-based fusion approach using multi-branch back-propagation neural networks to improve spectral mapping from high spatial resolution multispectral imagery to hyperspectral imagery. Their method leverages the assumption that spectral signatures are more similar within clusters than across clusters. However, this approach may introduce spectral distortion, leading to degraded performance in spectral angle mapper (SAM) metrics, especially when using bicubic decimation filters.
\par
\cite{wang2019deep} proposed an iterative fusion framework that jointly estimates the observation model and the fusion process in an alternating fashion. This blind fusion method enables the integration of low spatial resolution hyperspectral imagery and high spatial resolution multispectral imagery without prior knowledge of the observation model. The model includes an iterative refinement unit (IRU) and a dense fusion mechanism. However, the reliance on iterative back-projection introduces significant computational overhead, and the use of shared parameters in the IRU, while improving performance, may limit adaptability to diverse datasets.
\par
\cite{wang2022fsl} further introduced FSL-Unet, a U-Net–based model with spatial–spectral joint perceptual attention, designed to effectively integrate spatial information from high spatial resolution multispectral imagery and spectral information from low spatial resolution hyperspectral imagery. While effective in many cases, FSL-Unet occasionally struggles with noise and fine-detail distortions, resulting in suboptimal fusion performance.

\par
Physics-model–based approaches incorporate prior knowledge of the hyperspectral imaging process into deep learning frameworks to improve reconstruction fidelity. \cite{xie2020mhf} proposed MHF-Net, a multi-level hybrid fusion network that integrates spatial and spectral information through parallel convolutional branches and progressive fusion modules. By embedding the spectral response function into the network design, MHF-Net effectively constrains the reconstruction process to maintain spectral accuracy. \cite{wang2019deep} introduced DBIN, a deep blind image fusion network that jointly estimates the observation model and the fusion process without requiring prior alignment, making it robust to registration errors between hyperspectral and multispectral data. However, its iterative back-projection scheme introduces additional computational costs. \cite{dong2021model} presented MoG-DCN, a model-guided deformable convolution network that explicitly incorporates the physical image formation model while using deformable convolutions to adaptively capture geometric variations. This physics-guided design enhances performance in scenarios with limited training data, though it may require accurate knowledge of the sensor characteristics for optimal results.
\par
Transformer-based architectures have recently gained attention in hyperspectral image reconstruction for their ability to model long-range dependencies in both spatial and spectral dimensions. \cite{hu2022fusformer} proposed Fusformer, which employs multi-scale window-based attention to capture local and global spectral-spatial correlations, improving generalization to complex scenes. \cite{jia2023multiscale} introduced MSST-Net, a multi-scale spectral–spatial transformer network that combines convolutional layers and transformer blocks to efficiently model both short- and long-range dependencies across spectral bands. The SpecSwin3D model draws inspiration from the SwinUNETR architecture \cite{hatamizadeh2021swin}, which was originally designed for 3D brain tumor semantic segmentation. In SwinUNETR \cite{hatamizadeh2021swin}, the input consists of a 3D brain MRI image composed of multiple slices, and the output is a segmentation mask indicating the tumor. The model effectively captures both spatial and inter-slice dependencies by modeling not only the correlations within individual slices but also the relationships across adjacent slices. The Swin Transformer architecture excels in leveraging information from neighboring slices, which enhances the segmentation performance \cite{liu2021swin}.
\par
Beyond CNN-, physics model–, and Transformer-based approaches, other deep learning frameworks have also been explored for hyperspectral image reconstruction. In particular, generative adversarial networks (GANs) have been employed to enhance perceptual quality and reduce spectral distortion. GAN-based methods, such as those proposed in \cite{ren2020deep,qu2018udas}, utilize adversarial training to encourage the generated hyperspectral bands to match the distribution of real hyperspectral data, often in combination with auxiliary spectral consistency losses. These adversarial frameworks represent a promising direction for HIG, offering improved generalization and fidelity over conventional architectures, though they typically require larger datasets and more complex training procedures.

\section{Methods}\label{Methods}

\subsection{Study Areas}\label{Study_Areas}
\begin{figure}[ht!]
\begin{center}
		\includegraphics[width=1.0\columnwidth]{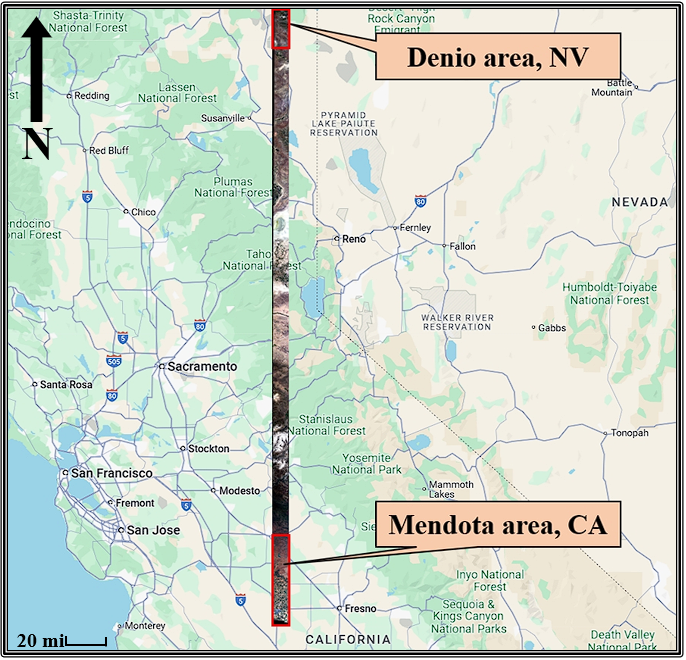}
    \caption{Study area corresponding to the AVIRIS site \textit{Bay Area Box Line 11}, which spans from northwest to southeast under clear-sky conditions. The dataset comprises orthorectified hyperspectral imagery with high spatial and spectral resolution, facilitating environmental analysis across diverse terrain types.}
\label{fig:study_area}
\end{center}
\end{figure}

This study utilizes a comprehensive hyperspectral dataset collected from California and Nevada (\autoref{fig:study_area}). The dataset is sourced from the publicly available AVIRIS (Airborne Visible/Infrared Imaging Spectrometer) repository \cite{aviris2025}. It consists of 3,000 orthorectified hyperspectral images with a spatial resolution of 14 meters and 224 spectral bands covering wavelengths from 400 nm to 2500 nm.

\subsection{Data Preprocessing}
\label{sec:data_preprocessing}
To ensure comparability across platforms, multispectral bands can generally be synthesized from hyperspectral imagery using spectral response functions (SRFs). For each target multispectral band, the linear combination coefficients $c_{ij}$ between the SRF of the target band and hyperspectral profiles are obtained by solving a least-squares problem \cite{blonski2003synthesis}:
\begin{equation}
R_i^{\mathrm{MSI}}(\lambda_k) = \sum_{j=1}^{N_{\mathrm{HSI}}} c_{ij} R_j^{\mathrm{HSI}}(\lambda_k), 
\quad k=1,\dots,n, \quad i=1,\dots,N_{\mathrm{MSI}},
\end{equation}
and the radiance values are synthesized as:
\begin{equation}
L_i^{\mathrm{MSI}} = \frac{\sum_{j=1}^{N_{\mathrm{HSI}}} c_{ij} \, \Delta_j \, L_j^{\mathrm{HSI}}}
{\sum_{j=1}^{N_{\mathrm{HSI}}} c_{ij} \, \Delta_j}.
\end{equation}
\par
\begin{figure}[ht!]
\begin{center}
		\includegraphics[width=0.7\columnwidth]{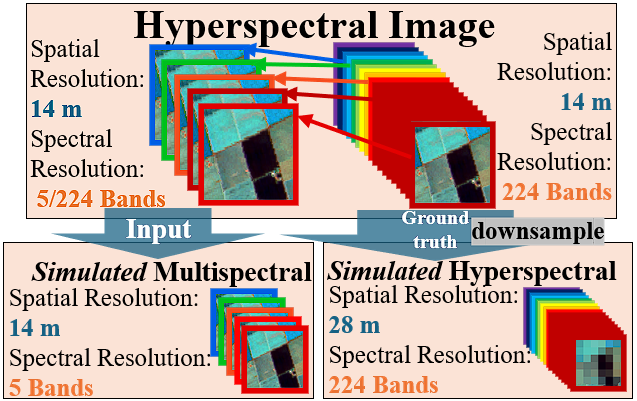}
	\caption{Simulation of Hyperspectral and Multispectral Bands Using Hyperspectral Imagery: Five hyperspectral bands were selected to match the wavelengths of multispectral bands, while all hyperspectral bands were downsampled to simulate lower spatial resolution hyperspectral bands.}
\label{fig:data_preprocessing}
\end{center}
\end{figure}

In this study, since both the multispectral and hyperspectral data are derived from the AVIRIS dataset, we directly selected five hyperspectral bands to serve as simulated high-spatial-resolution (HSR) multispectral inputs (\autoref{fig:data_preprocessing}), without requiring additional SRF-based synthesis.
\par
Subsequently, we downsampled the spatial resolution of the hyperspectral image from 14 m to 28 m via bilinear interpolation to simulate lower spatial resolutions, which were treated as simulated low-spatial-resolution (LSR) hyperspectral bands. In essence, the original spatial resolution hyperspectral bands were used as multispectral bands, while the downsampled hyperspectral image served as quasi-hyperspectral bands.
\par
During training stage, each image was cropped into 128 × 128 pixel tiles, which served as the model’s input. To enhance the robustness of the model, data augmentation techniques were applied, including random rotations within a range of -15$^\circ$ to 15$^\circ$, random cropping at 70\% to 90\% of the original image size, and horizontal and vertical flipping. The dataset was partitioned into three subsets: training (70\%), validation (15\%), and test (15\%), using a stratified split to maintain class distribution.
\par
For model training, the downsampled hyperspectral bands were used as the ground truth, while the simulated multispectral bands served as the input. The model was trained to learn the relationships between the multispectral bands and their corresponding hyperspectral bands. After training, the model was tested using the simulated multispectral bands to generate hyperspectral bands. The generated hyperspectral bands were then evaluated against the downsampled hyperspectral ground truth to assess the model’s performance.
\par
Specifically, we selected five representative hyperspectral bands—band 9 (459 nm, blue), band 20 (553 nm, green), band 30 (672 nm, red), band 40 (846 nm, near-infrared), and band 52 (1240 nm, shortwave infrared)—to serve as simulated multispectral inputs.
\subsection{Model Architecture}
\label{sec:Model}

\subsubsection{Model Overview}
SpecSwin3D is a 3D Transformer–based encoder–decoder framework designed for hyperspectral image generation. It models both local and long-range dependencies by capturing correlations within each spectral slice and across adjacent slices, inspired by advances in volumetric imaging architectures \cite{liu2021swin,hatamizadeh2021swin}.
\par
The key idea is to reproject spectral bands into the depth dimension of a 3D volume, allowing the model to treat neighboring wavelengths as adjacent slices. Just as adjacent slices in volumetric imaging often share strong correlations, hyperspectral bands exhibit high continuity across nearby wavelengths \cite{chang2000information}. By exploiting this property with 3D shifted-window attention, SpecSwin3D enhances reconstruction fidelity—especially for spectrally adjacent bands—while preserving fine spatial detail.
\par
Our approach to constructing the 3D volume is not a simple sequential stacking of bands, but a deliberately designed sequence that accounts for inter-slice dependencies; the detailed design of this sequence is presented in Section~\ref{sec:sequence}.

\begin{figure}[ht!]
\begin{center}
		\includegraphics[width=1\columnwidth]{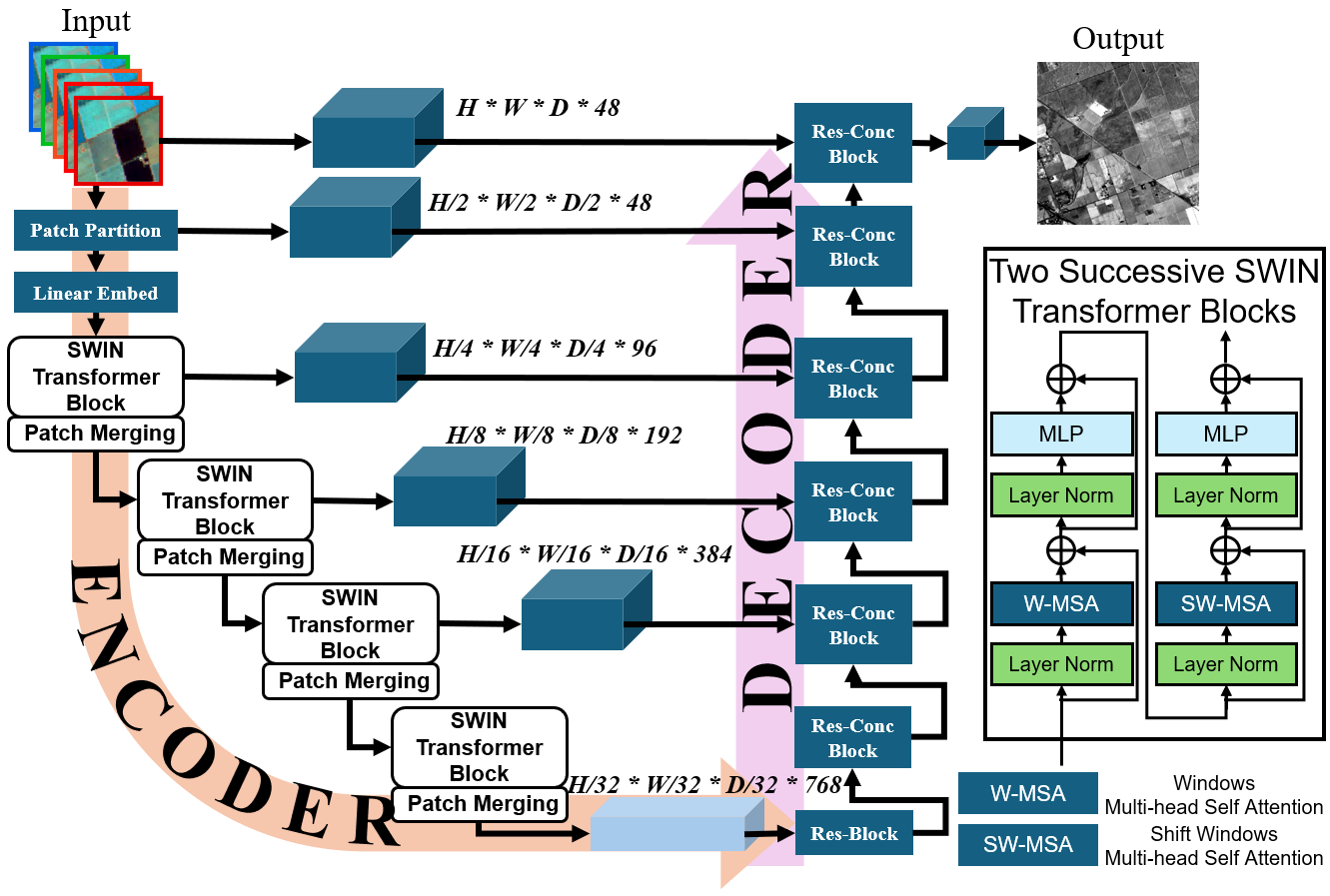}
	\caption{Overview of the SpecSwin3D architecture: The input consists of multispectral images with 5 bands, which are first divided into non-overlapping patches and embedded for efficient 3D self-attention. Encoded features are progressively reconstructed through a multi-scale hierarchical decoder with skip connections across different resolutions. The final output can be either a single band or multiple bands, depending on the reconstruction task.}
\label{fig:architecture}
\end{center}
\end{figure}

\subsubsection{Swin3D Encoder Details}
The SpecSwin3D architecture integrates the Swin Transformer encoder \cite{liu2021swin} within a U-Net framework to effectively capture hierarchical information from the input data. As depicted in \autoref{fig:architecture}, the model’s input is first processed through a patch partitioning and linear embedding block, which transforms the image into 3D tokens. In the encoding stage, two successive Swin Transformer blocks are employed to efficiently compute self-attention with reduced computational complexity. The self-attention mechanism is defined as:

\begin{equation}\label{eq:QKV}
\text{Attention}(Q, K, V) = \text{Softmax}\left(\frac{QK^T}{\sqrt{d_k}}\right) V,
\end{equation}

\begin{tabbing} 
where \\
\hspace{0.6cm} \= $Q$ = query matrix\\
\> $K$ = key matrix\\
\> $V$ = value matrix\\
\> $d_k$ = dimension of the key matrix
\end{tabbing}

The Window Multi-Head Self-Attention (W-MSA) blocks utilize non-overlapping windows to compute self-attention within each window. To enhance information exchange between different windows, the Shifted Window Multi-Head Self-Attention (SW-MSA) mechanism is applied. Specifically, SW-MSA introduces a 3D cyclic-shifting window that facilitates the flow of information across adjacent windows, with particular emphasis on capturing relationships across different channels. Following the SWIN Transformer blocks, the patch merging layers reduce the spatial dimensions (height and width) while increasing the channel dimension through linear transformations. This downsampling mechanism not only decreases the spatial resolution but also enriches the model’s capacity to capture long-range dependencies by utilizing more channels. This encoding phase significantly contributes to expanding the global receptive field in an efficient manner.

\subsubsection{Decoder and Output Strategy}
In the decoder stage, the model adopts a U-Net architecture for upsampling sequences at various resolutions. Residual blocks are employed to upsample intermediate feature maps, while skip connections are used to merge information across different scales. This fusion of local and global features enhances the model’s performance in HIG tasks.

To supervise the HIG result, we adopt the root mean squared error (RMSE) as the loss function, computed across all target bands. This objective encourages the model to minimize reconstruction errors uniformly over the entire spectral range. Specifically, the loss is defined as:

\begin{equation}
\mathcal{L}_{\mathrm{RMSE}} = \frac{1}{B \cdot C \cdot H \cdot W} \sum_{b=1}^{B} \sum_{c=1}^{C} \sum_{h=1}^{H} \sum_{w=1}^{W} \left( \hat{y}_{bchw} - y_{bchw} \right)^2
\end{equation}

\begin{tabbing}
Where: \\
\hspace{0.6cm} \= $B$ \= is the batch size. \\
\> $C$ \= is the number of target spectral bands. \\
\> $H$ \= and $W$ \= are the height and width of each image tile. \\
\> $\hat{y}_{bchw}$ \= and $y_{bchw}$ \= represent the predicted and ground truth\\
\> pixel values at batch index $b$, band index $c$, and spatial\\
\> position $(h, w)$.
\end{tabbing}

By minimizing this loss, the model is guided to reduce the spectral reconstruction error across all target bands, ensuring balanced learning without favoring specific wavelengths.

\subsubsection{Band Sequence Design for 3D Shifted Window}
\label{sec:sequence}
To effectively model inter-band relationships within the multispectral inputs (\textit{bands} [9, 20, 30, 40, 52], as defined in Section~\ref{sec:data_preprocessing}), we designed a 3D input of shape \(H \times W \times D \times S\), where \(H\) and \(W\) denote spatial dimensions, \(D\) is the depth (i.e., number of spectral slices), and \(S\) is the number of channels per slice. In our implementation, the depth dimension is used to organize the sequence of spectral bands, allowing the SpecSwin3D's SW-MSA mechanism to capture 3D context across adjacent spectral slices.

\begin{figure}[ht!]
\begin{center}
\includegraphics[width=0.7\columnwidth]{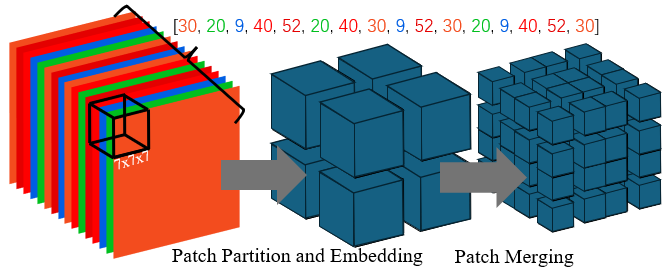}
\caption{Overview of shifted window, patch partition, embedding, and patch merging.}
\label{fig:token}
\end{center}
\end{figure}

To guarantee that all \(\binom{5}{2} = 10\) unique band pairs among the five selected input bands co-occur within at least one adjacent slice pair in the depth dimension, we designed a compact input sequence of length \(D = 16\):
[30, 20, 9, 40, 52, 20, 40, 30, 9, 52, 30, 20, 9, 40, 52, 30]. This sequence ensures complete pairwise coverage.

Formally, let \(d_i\) and \(d_j\) denote the indices (depth positions) of bands \(i\) and \(j\) within the input sequence, respectively. We define adjacent co-occurrence as \(|d_i - d_j| = 1\), indicating that the two bands are placed in consecutive depth slices. This constraint ensures that every band pair is locally accessible to the self-attention mechanism operating within the shifted 3D windows. Longer-range spectral relationships, beyond immediate adjacency, are expected to be captured by the global modeling capacity of the transformer backbone.

\subsection{Cascade Training Strategy and Fine-tuning Approach}
\label{sec:cascade}

\subsubsection{Overview}
In our research, our goal is to use 5 multispectral bands to generate 224 hyperspectral bands. There are two modeling approaches. The first approach involves training 224 separate models, where each model generates one corresponding hyperspectral band. The second approach uses just one model to generate all 224 bands simultaneously. The first method can be conceptualized as outputting 224 single-channel images, while the second produces one image with 224 channels.

There is a significant trade-off between these two training strategies. Training one model to output all 224 bands performs poorly because the model optimizes for the average loss across all bands. This means that for individual bands, the gradient descent direction is often suboptimal. In contrast, training 224 separate models ensures that each band follows its optimal gradient descent path. However, this approach requires substantially more computational resources and time.

\subsubsection{Stage 1: Cascade Training}
To balance performance and computational resources, we implemented cascade training strategies combined with fine-tuning. First, we classified all 224 bands into 6 classes using various classification methods. Classes 1 through 5 comprise cascade bands—the 29 selected bands trained in the first stage. 
\par
\begin{figure}[ht!]
\begin{center}
		\includegraphics[width=1\columnwidth]{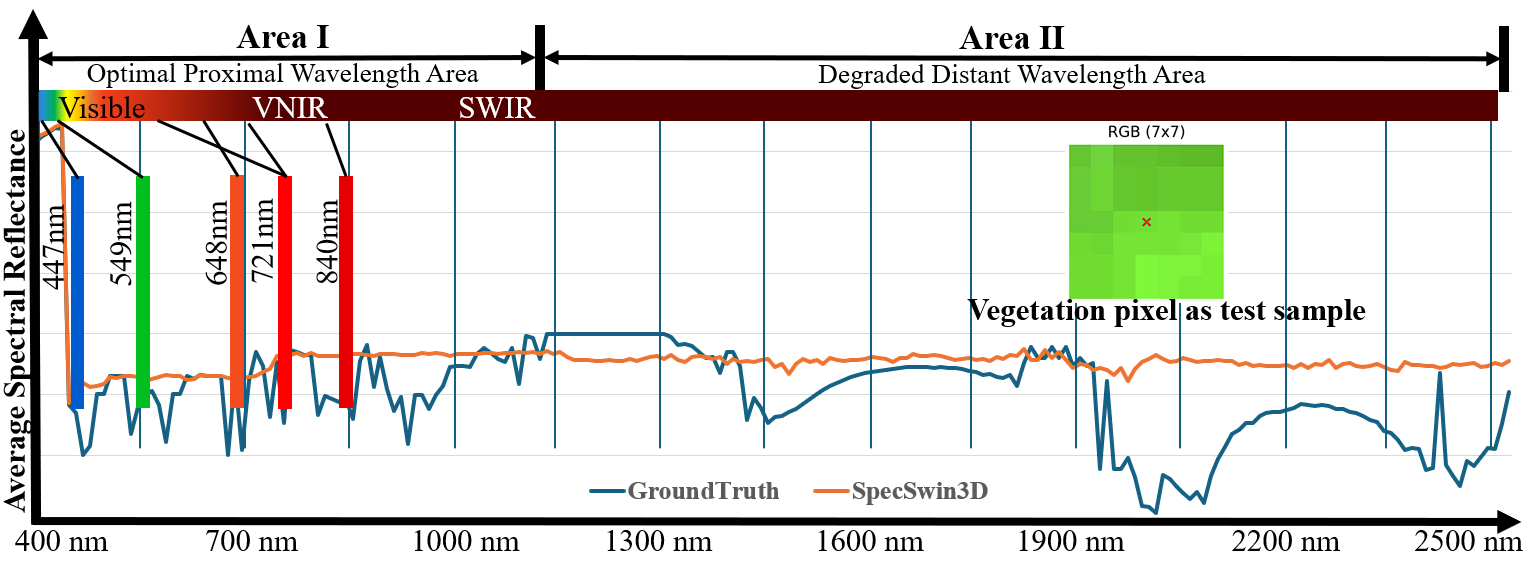}
	\caption{Spectral reconstruction performance for a vegetation pixel, selected as the representative test case (inset RGB patch). The orange curve denotes SpecSwin3D predictions and the blue curve denotes the ground truth. Similar to \autoref{fig:preliminary_result}, two wavelength regions are highlighted: Area I (proximal bands with high accuracy) and Area II (distant bands with reduced accuracy). Note that the y-axis is log-scaled to fit the full spectral range; this amplifies visual deviations, while absolute reconstruction errors remain small.}
\label{fig:band_region_overall}
\end{center}
\end{figure}

In \autoref{fig:band_region_overall}, the results indicate that the HIG accuracy varies with spectral distance from the input multispectral bands. Therefore, defining an appropriate cascade pyramid structure is essential. We generated different cascade pyramids using various cascade strategies, with the specific hierarchical groupings summarized in Table~\ref{tab:cascade_strategy}. 
\par
During training, we progressively expanded the spectral range across pyramid levels, with the number of epochs decaying by a factor of 0.9 at each level while enforcing a minimum of 40 epochs. The detailed epoch scheduling for each cascade level is summarized in Table~\ref{tab:cascade_epochs}.

\begin{table}[t]
\centering
\caption{Epoch settings for cascade training. A decay factor of 0.9 was applied at each level, with a minimum of 40 epochs enforced. The fine-tuning stage further adjusts epochs based on the spectral distance factor (\textit{sd\_factor}).}
\label{tab:cascade_strategy}
\scriptsize
\renewcommand{\arraystretch}{1.2}

\begin{tabularx}{\textwidth}{lXXXXX}
\toprule
\textbf{Strategy} & \textbf{Level 0} & \textbf{Level 1} & \textbf{Level 2} & \textbf{Level 3} & \textbf{Level 4} \\
\midrule
Physical & [0, 1, 2, 3, 4, 5, 6, 7, 8] & [15, 25, 27] & [48, 50, 54, 67, 81, 83, 90] & [108, 125, 135] & [155, 162, 175, 185, 189, 210, 218] \\
Mutual Info & [9, 18, 26, 27, 28, 29, 30, 31, 32] & [16, 19, 25] & [8, 10, 14, 15, 17, 20, 24] & [11, 12, 13] & [7, 21, 22, 23, 33, 47, 48] \\
Variance & [115, 165, 166, 167, 168, 169, 170, 171, 172] & [114, 116, 173] & [111, 112, 113, 117, 118, 119, 174] & [164, 175, 176] & [120, 177, 178, 207, 208, 209, 211] \\
Spectral Physics & [70, 71, 72, 80, 81, 82, 83, 84, 103] & [73, 74, 75] & [76, 77, 78, 79, 85, 86, 87] & [88, 89, 90] & [91, 92, 93, 95, 96, 97, 98] \\
Correlation & [16, 19, 20, 21, 22, 23, 24, 25, 35] & [15, 17, 18] & [12, 13, 14, 26, 27, 28, 29] & [11, 30, 31] & [8, 9, 10, 32, 33, 34, 36] \\
Uniform & [0, 7, 15, 23, 31, 38, 46, 54, 62] & [70, 77, 85] & [93, 101, 109, 116, 124, 132, 140] & [147, 155, 163] & [171, 179, 186, 194, 202, 210, 218] \\
\bottomrule
\end{tabularx}

\vspace{0.5em}
\raggedright
\footnotesize
\textbf{Physical} – Based on physical spectral characteristics, ordered distribution from blue light to infrared.  
\textbf{Mutual Info} – Based on mutual information importance, concentrated in visible light and near-infrared.  
\textbf{Variance} – Based on variance importance, mainly in infrared and near-infrared regions.  
\textbf{Spectral Physics} – Based on spectral physical properties, concentrated on specific absorption peaks.  
\textbf{Correlation} – Based on correlation analysis, bands are more dispersed.  
\textbf{Uniform} – Mathematical uniform sampling.
\end{table}

\begin{table}[t]
\centering
\caption{Epoch Settings for Cascade Training}
\label{tab:cascade_epochs}
\small
\begin{tabular}{|l|c|c|c|}
\hline
\textbf{Level} & \begin{tabular}{@{}c@{}}\textbf{Epoch} \\ \textbf{Factor}\end{tabular} &
\begin{tabular}{@{}c@{}}\textbf{Calculation} \\ \textbf{Formula}\end{tabular} &
\begin{tabular}{@{}c@{}}\textbf{Actual} \\ \textbf{Epochs}\end{tabular} \\
\hline
Level 0 & $0.9^0$ & $\max(40,\ 80 \times 1.0)$ & 80 \\
Level 1 & $0.9^1$ & $\max(40,\ 80 \times 0.9)$ & 72 \\
Level 2 & $0.9^2$ & $\max(40,\ 80 \times 0.81)$ & 65 \\
Level 3 & $0.9^3$ & $\max(40,\ 80 \times 0.73)$ & 58 \\
Level 4 & $0.9^4$ & $\max(40,\ 80 \times 0.66)$ & 53 \\
Fine-tuning & \textit{sd\_factor} & \textit{basic\_epoch} $\times$ \textit{sd\_factor} & 21--104 \\
\hline
\end{tabular}
\end{table}

\subsubsection{Stage 2: Fine-tuning Approach}
Class 6 consists of bands fine-tuned in the second stage based on the cascade bands. We determined the fine-tuning epochs for class 6 bands according to similarity scores between different bands and their spectral distances. The detailed settings are summarized in Table~\ref{tab:finetuning_epochs}.

\begin{table}[t]
\centering
\caption{Epoch settings for fine-tuning bands (Class 6). Fine-tuning epochs are adjusted according to band similarity scores and spectral distances. A base epoch value is scaled by a distance factor (0.7, 1.0, or 1.3) depending on the spectral distance.}
\label{tab:finetuning_epochs}
\small
\begin{tabular}{|c|c|c|c|c|}
\hline
\begin{tabular}{@{}c@{}}\textbf{Similarity} \\ \textbf{Score}\end{tabular} &
\begin{tabular}{@{}c@{}}\textbf{Band} \\ \textbf{Distance}\end{tabular} &
\begin{tabular}{@{}c@{}}\textbf{Base} \\ \textbf{Epochs}\end{tabular} &
\begin{tabular}{@{}c@{}}\textbf{Distance} \\ \textbf{Factor}\end{tabular} &
\begin{tabular}{@{}c@{}}\textbf{Actual} \\ \textbf{Epochs}\end{tabular} \\
\hline
$> 0.8$ (High)     & $< 5$     & 30 & $\times 0.7$     & 21  \\
$> 0.8$ (High)     & 5--50     & 30 & --               & 30  \\
$> 0.8$ (High)     & $> 50$    & 30 & $\times 1.3$     & 39  \\
0.6--0.8 (Medium)  & $< 5$     & 50 & $\times 0.7$     & 35  \\
0.6--0.8 (Medium)  & 5--50     & 50 & --               & 50  \\
0.6--0.8 (Medium)  & $> 50$    & 50 & $\times 1.3$     & 65  \\
$< 0.6$ (Low)      & $< 5$     & 80 & $\times 0.7$     & 56  \\
$< 0.6$ (Low)      & 5--50     & 80 & --               & 80  \\
$< 0.6$ (Low)      & $> 50$    & 80 & $\times 1.3$     & 104 \\
\hline
\end{tabular}
\end{table}

\subsubsection{Summary}
By adopting this two-stage cascade and fine-tuning strategy, our model significantly reduces training time while maintaining high HIG performance. This design enables a favorable trade-off between computational efficiency and spectral fidelity.

\subsection{Evaluation Metrics}
To evaluate the effectiveness of our approach, we compared it against five commonly used HIG deep learning methods. The baseline methods selected for benchmarking include MHF-Net \cite{xie2020mhf}, DBIN \cite{wang2019deep}, Fusformer \cite{hu2022fusformer}, MoG-DCN \cite{dong2021model}, and MSST-Net \cite{jia2023multiscale}.
\par
Five widely used metrics, including Peak Signal-to-Noise Ratio (PSNR)\cite{box1988signal}, Erreur Relative Globale Adimensionnelle de Synthèse (ERGAS)\cite{wald2002data,wald2000quality}, Spectral Angle Mapper (SAM)\cite{girouard2004validated,boardman1993automating}, Q Index\cite{wang2002universal}, and Structural Similarity Index Metric (SSIM)\cite{wang2004image}, are employed to evaluate the quality of the constructed hyperspectral bands comprehensively.

\textbf{PSNR} is a common metric for evaluating the noise level or distortion in reconstructed images. A higher PSNR value indicates less distortion and better image quality. It is given by:

\begin{equation}
\text{PSNR} = 10 \log_{10}\!\left(\frac{\max(f)^2}{\text{MSE}}\right)
\end{equation}

\begin{equation}
\text{MSE} = \frac{1}{MNK} \sum_{k=1}^{K} \sum_{i=1}^{M} \sum_{j=1}^{N} \bigl(f_k(i,j) - \hat{f}_k(i,j)\bigr)^2
\end{equation}

\begin{tabbing}
Where:\\
\hspace{0.6cm} \= $\text{max}(f)$ \= is the maximum pixel value of the reference image. \\
\> $f_k(i,j)$ and $\hat{f_k}(i,j)$ \= are the pixel values at position $(i,j)$ for \\
\> the $k$-th band of the reference and reconstructed images, \\
\> respectively. $M$, $N$, and $K$ \= are the dimensions \\
\> of the hyperspectral image.
\end{tabbing}

\textbf{ERGAS} evaluates the overall quality of the fused image, accounting for both spectral and spatial distortions. Lower ERGAS values indicate better fusion results. It is given by:

\begin{equation}
\text{ERGAS} = 100 \cdot \left( \frac{1}{B} \sum_{b=1}^{B} \left( \frac{\text{RMSE}_b}{\mu_b} \right)^2 \right)^{0.5}
\end{equation}

\begin{tabbing}
Where: \\
\hspace{0.6cm}  \= $B$ \= is the number of spectral bands. \\
\> $\text{RMSE}_b$ \= and $\mu_b$ \= are the root mean square error and\\ 
\> mean value of the $b$-th band, respectively.
\end{tabbing}

\textbf{SAM} measures the spectral similarity between the reference and reconstructed images by computing the averaged spectral angle across all pixels. Smaller SAM values indicate better spectral reconstruction. It is given by:

\begin{equation}
\text{SAM} = \frac{1}{MN} \sum_{i=1}^{M} \sum_{j=1}^{N} \arccos \left( \frac{\langle f(i,j), \hat{f}(i,j) \rangle}{\| f(i,j) \| \| \hat{f}(i,j) \|} \right)
\end{equation}

\begin{tabbing}
Where:\\
\hspace{0.6cm} \= $f(i,j)$ \= and $\hat{f}(i,j)$ \= \\
\> are spectral vectors of the reference and reconstructed \\
\> images at pixel $(i,j)$.\\
\> $\langle \cdot, \cdot \rangle$ \= denotes the dot product, \= and $\| \cdot \|$ \= represents the\\
\> vector norm.
\end{tabbing}

The \textbf{Q index} quantifies the similarity between the reference and reconstructed images by considering the correlation, luminance, and contrast of each band. Higher Q values indicate better reconstruction quality. It is given by:

\begin{equation}
\text{Q} = \frac{1}{B} \sum_{b=1}^{B} \frac{(2\mu_b \hat{\mu}_b + C_1) (2\sigma_{b,b} + C_2)}{\left( \mu_b^2 + \hat{\mu}_b^2 + C_1 \right) \left( \sigma_b^2 + \hat{\sigma}_b^2 + C_2 \right)}
\end{equation}

\begin{tabbing}
Where: \\
\hspace{0.6cm} \= $\mu_b$ \= and $\hat{\mu}_b$ \= are the mean intensities of the $b$-th band. \\
\> $\sigma_b^2$, $\hat{\sigma}_b^2$, \= and $\sigma_{b,b}$ \= are the variances and covariance. \\
\> $C_1$ \= and $C_2$ \= are constants.
\end{tabbing}

\textbf{SSIM} evaluates the similarity in structural information between the reference and reconstructed images. Higher SSIM values imply better spatial structure preservation. It is given by:

\begin{equation}
\text{SSIM} = \frac{1}{K} \sum_{k=1}^{K} \frac{(2\mu_k \hat{\mu}_k + C_1)(2\sigma_{k,k} + C_2)}{(\mu_k^2 + \hat{\mu}_k^2 + C_1)(\sigma_k^2 + \hat{\sigma}_k^2 + C_2)}
\end{equation}

\begin{tabbing}
Where: \\
\hspace{0.6cm} \=  $\mu_k$ \= and $\hat{\mu}_k$ \= are the mean intensities of the $k$-th band for\\
\> the reference and reconstructed images. \\
\> $\sigma_k^2$ \= and $\hat{\sigma}_k^2$ \= are the variances, \= and $\sigma_{k,k}$ \= is the covariance. \\
\> $C_1$ \= and $C_2$ \= are small constants to stabilize the division.
\end{tabbing}

\section{Results}
\label{sec:result}
\renewcommand{\arraystretch}{1.3} 

\begin{table}[h]
    \centering
    \begin{threeparttable}
        \caption{Quantitative Comparison of Different Models on Hyperspectral Reconstruction.}
        \begin{tabular}{|l|c|c|c|c|c|}\hline
            \textbf{Models} \rule{0pt}{1.1em} & \textbf{PSNR$\uparrow$} & \textbf{ERGAS$\downarrow$} & \textbf{SAM$\downarrow$} & \textbf{Q Index$\uparrow$} & \textbf{SSIM$\uparrow$} \\\hline
            MHF-Net     & 30.18 & 1.61 & 4.14 & 0.82 & 0.93 \\
            DBIN        & 31.91 & 1.54 & 3.37 & 0.89 & 0.93 \\
            Fusformer   & 32.14 & 1.57 & 3.05 & 0.88 & 0.94 \\
            MoG-DCN     & 33.34 & 0.80 & 3.18 & 0.93 & \textbf{0.96} \\
            MSST-Net    & 31.18 & 0.83 & 2.73 & 0.95 & 0.95 \\
            SpecSwin3D  & \textbf{35.82} & \textbf{0.77} & \textbf{2.40} & \textbf{0.97} & \textbf{0.96} \\\hline
        \end{tabular}
        \begin{tablenotes}
            \footnotesize
            \item All models were tested on 30 samples, and the results are averaged over the best 10 spectral bands.
        \end{tablenotes}
        \label{tab:model_performance}
    \end{threeparttable}
\end{table}

We quantitatively evaluated the performance of the proposed SpecSwin3D model against five state-of-the-art HIG methods, including MHF-Net \cite{xie2020mhf}, DBIN \cite{wang2019deep}, Fusformer \cite{hu2022fusformer}, MoG-DCN \cite{dong2021model}, and MSST-Net \cite{jia2023multiscale}. The evaluation was conducted using five standard image quality metrics: PSNR, ERGAS, SAM, Q Index, and SSIM. The results are summarized in \autoref{tab:model_performance}.
\par
In terms of \textbf{PSNR}, SpecSwin3D achieved the highest score of \textbf{35.82 dB}, which is \textbf{+2.48 dB} higher than the next best MoG-DCN (33.34 dB) and \textbf{+5.64 dB} higher than the baseline MHF-Net (30.18 dB). This indicates that SpecSwin3D preserves pixel-level fidelity and suppresses noise more effectively than both the strongest competing method and the classical baseline.
\par
For \textbf{ERGAS}, which measures global reconstruction quality with lower values indicating better performance, SpecSwin3D obtained the lowest score of \textbf{0.77}. This slightly outperforms MoG-DCN (0.80) and MSST-Net (0.83), while significantly surpassing Fusformer (1.57), DBIN (1.54), and MHF-Net (1.61). These results demonstrate SpecSwin3D's robustness in preserving spatial and spectral consistency.
\par
Regarding \textbf{SAM}, SpecSwin3D achieved the best performance with the lowest score of \textbf{2.40°}. It outperformed MSST-Net (2.73°), Fusformer (3.05°), MoG-DCN (3.18°), DBIN (3.37°), and MHF-Net (4.14°), confirming its superior ability to maintain spectral fidelity.
\par
For \textbf{Q Index}, which evaluates the overall similarity between reconstructed and reference images, SpecSwin3D reached the highest score of \textbf{0.97}, outperforming MSST-Net (0.95) and MoG-DCN (0.93). This result further confirms its strength in capturing both spectral and structural features.
\par
Finally, for \textbf{SSIM}, SpecSwin3D tied for the best performance with MoG-DCN, both achieving \textbf{0.96}, while MSST-Net followed at 0.95 and the remaining models scored at or below 0.94. This suggests that SpecSwin3D effectively preserves spatial structure while maintaining overall spectral fidelity.

\begin{figure}[htbp!]
\centering
\begin{subfigure}[b]{\textwidth}
    \centering
    \includegraphics[width=0.9\textwidth]{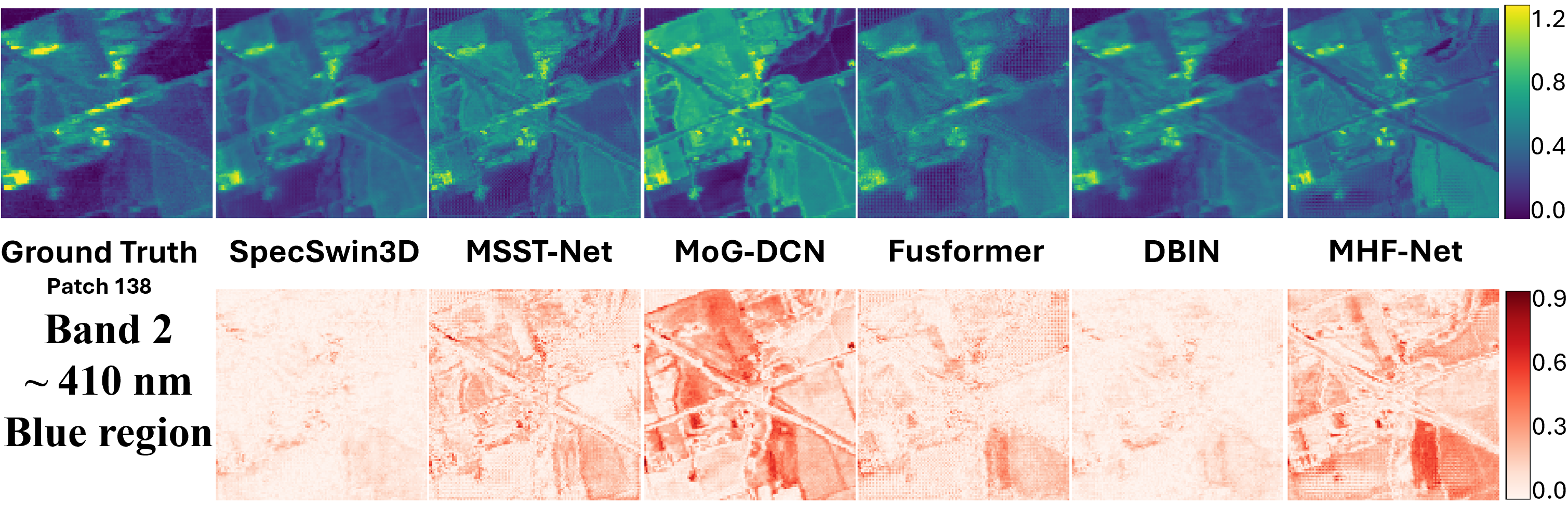}
    \caption{Band 2}
    \label{fig:result_band2}
\end{subfigure}

\begin{subfigure}[b]{\textwidth}
    \centering
    \includegraphics[width=0.9\textwidth]{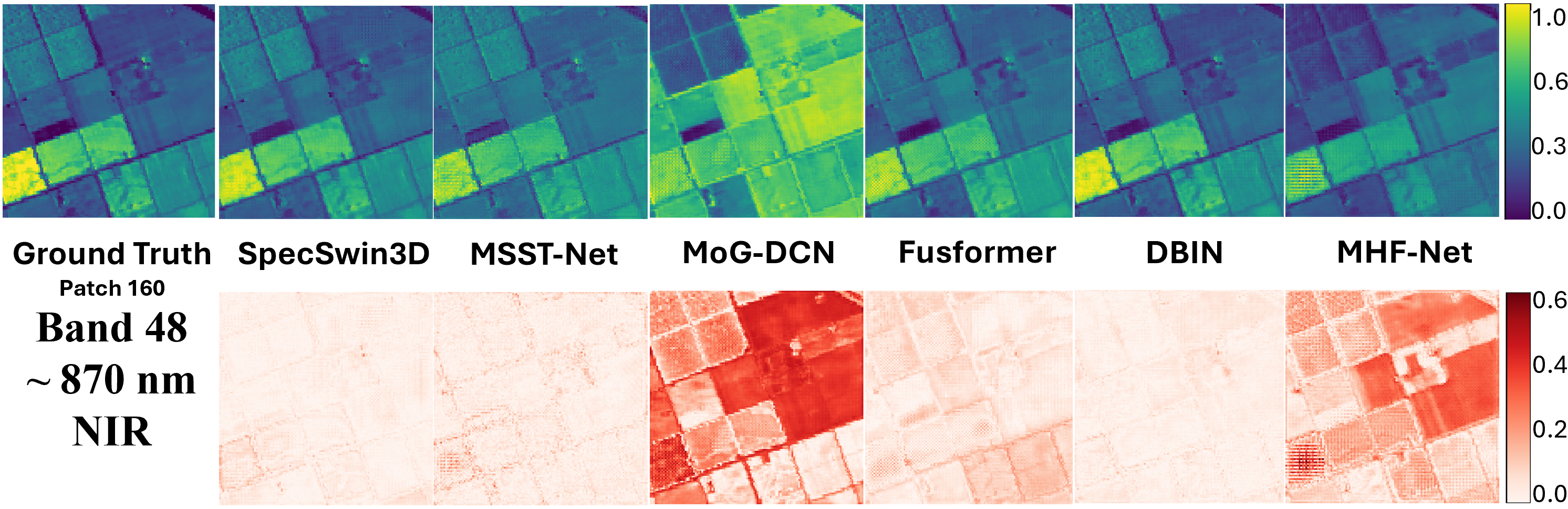}
    \caption{Band 48}
    \label{fig:result_band48}
\end{subfigure}

\begin{subfigure}[b]{\textwidth}
    \centering
    \includegraphics[width=0.9\textwidth]{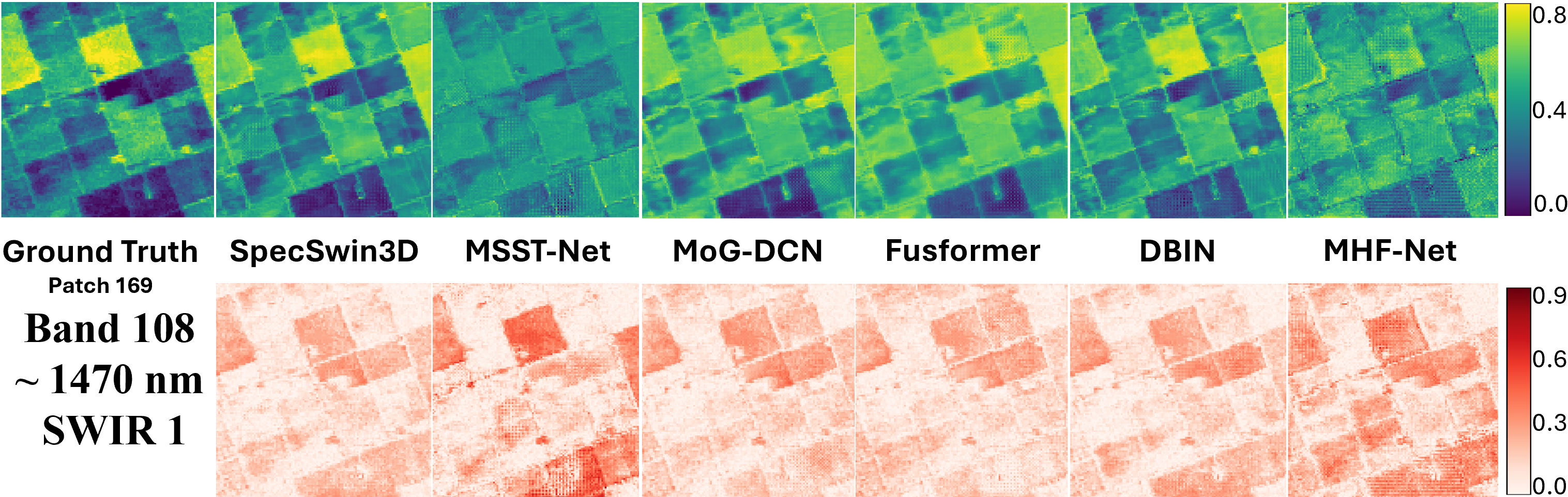}
    \caption{Band 108}
    \label{fig:result_band108}
\end{subfigure}

\begin{subfigure}[b]{\textwidth}
    \centering
    \includegraphics[width=0.9\textwidth]{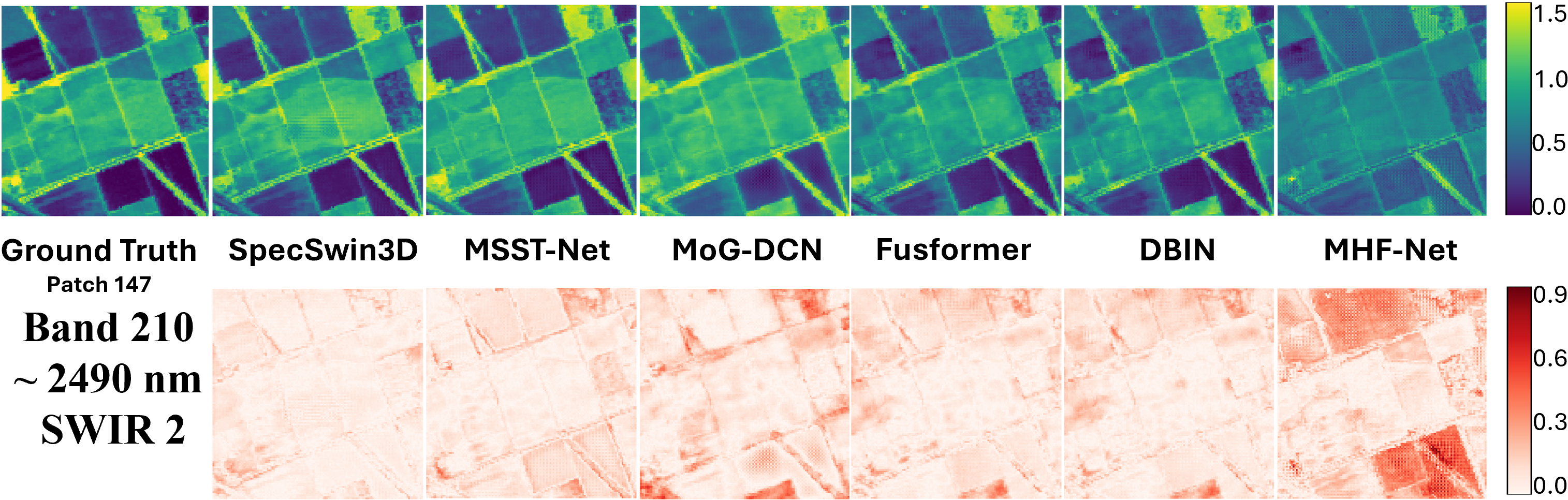}
    \caption{Band 210}
    \label{fig:result_band210}
\end{subfigure}

\caption{Model performance for selected band reconstruction across different spectral ranges. SpecSwin3D achieves state-of-the-art performance, consistently surpassing competing methods in all metrics and preserving spatial structures and spectral fidelity across the entire spectral range.}
\label{fig:result}
\end{figure}

\autoref{fig:result} illustrates the qualitative evaluation of the proposed SpecSwin3D model by visualizing generated results for four representative hyperspectral bands: (\subref{fig:result_band2}) Band~2 (410~nm, blue region), (\subref{fig:result_band48}) Band~48 (870~nm, NIR region), (\subref{fig:result_band108}) Band~108 (1470~nm, SWIR-1 region), and (\subref{fig:result_band210}) Band~210 (2490~nm, SWIR-2 region), extracted from a selected \(128 \times 128\) image tile. These visual examples facilitate direct inspection of the model’s ability to recover both spatial structures and spectral details.

\par
Among the evaluated methods, SpecSwin3D, MSST-Net, and MoG-DCN demonstrate satisfactory performance, which aligns with their relatively high quantitative scores across the five evaluation metrics reported in \autoref{tab:model_performance}. 
\par
\textbf{Band 2 (RGB)\autoref{fig:result_band2}.} SpecSwin3D produces the reconstruction most consistent with the ground truth, exhibiting minimal noise and the highest overall fidelity. MSST\mbox{-}Net performs reasonably in global appearance but suffers from clear edge distortions in the top\mbox{-}left region and introduces considerable noise. MoG\mbox{-}DCN shows poor reconstruction quality, particularly in the top\mbox{-}left and bottom\mbox{-}right regions. Fusformer delivers weak overall quality, whereas DBIN is passable. MHF\mbox{-}Net is generally inferior, with especially poor performance in the bottom\mbox{-}right region, although its edge recovery is relatively better.
\par
\textbf{Band 48 (NIR)\autoref{fig:result_band48}.} SpecSwin3D achieves the best overall reconstruction, with significantly reduced noise; only the field boundary in the bottom\mbox{-}right region appears less accurate. MSST\mbox{-}Net is generally acceptable but introduces substantial noise. MoG\mbox{-}DCN performs worst in the NIR band, even though its metrics appear favorable in \autoref{tab:model_performance} because the table aggregates results across all spectral regions. Fusformer also produces noisy and poor reconstructions, particularly in the bottom\mbox{-}right region. DBIN performs well overall, though it exhibits noise in the top\mbox{-}right region. MHF\mbox{-}Net is inferior overall, slightly better than MoG\mbox{-}DCN, but shows especially poor quality in the right\mbox{-}side farmland.
\par
\textbf{Band 108 (SWIR-1)\autoref{fig:result_band108}.} The input contains noticeable distortions, yet SpecSwin3D successfully identifies and corrects them, producing a reconstruction closest to the ground truth. MSST\mbox{-}Net performs poorly overall, with particularly weak contrast recovery. MoG\mbox{-}DCN, Fusformer, and DBIN deliver comparable results that are generally acceptable. In contrast, MHF\mbox{-}Net is clearly affected by the input distortions, leading to farmland boundaries that are no longer straight.
\par
\textbf{Band 210 (SWIR-2)\autoref{fig:result_band210}.} SpecSwin3D delivers the best overall performance, with only minor noise in the bottom\mbox{-}left and right regions and slight edge errors, yet it still surpasses other methods in the SWIR-2 region. MSST\mbox{-}Net and MoG\mbox{-}DCN are generally acceptable but show poor edge quality in the upper\mbox{-}middle area and noise in the bottom\mbox{-}right. Fusformer and DBIN also provide acceptable reconstructions, though noise is mainly concentrated in the upper and lower parts of the image. MHF\mbox{-}Net performs worst overall, with large errors in both the bottom\mbox{-}right and upper farmlands, as well as severe noise along the right side.
\par
In summary, SpecSwin3D consistently outperforms competing methods across all evaluation metrics and maintains superior reconstruction quality across all bands spanning the entire spectral range. This highlights its robustness in preserving both spatial structures and spectral fidelity throughout the RGB, NIR, and SWIR regions. It achieves the best results in all five metrics. These quantitative improvements, together with the qualitative reconstructions shown in \autoref{fig:result}, highlight the robustness and effectiveness of SpecSwin3D in generating high-quality hyperspectral imagery across a diverse spectral range.
\section{Discussion}
\subsection{Band Sequence Design}
\begin{figure}[H]
    \centering
    \includegraphics[width=0.9\columnwidth]{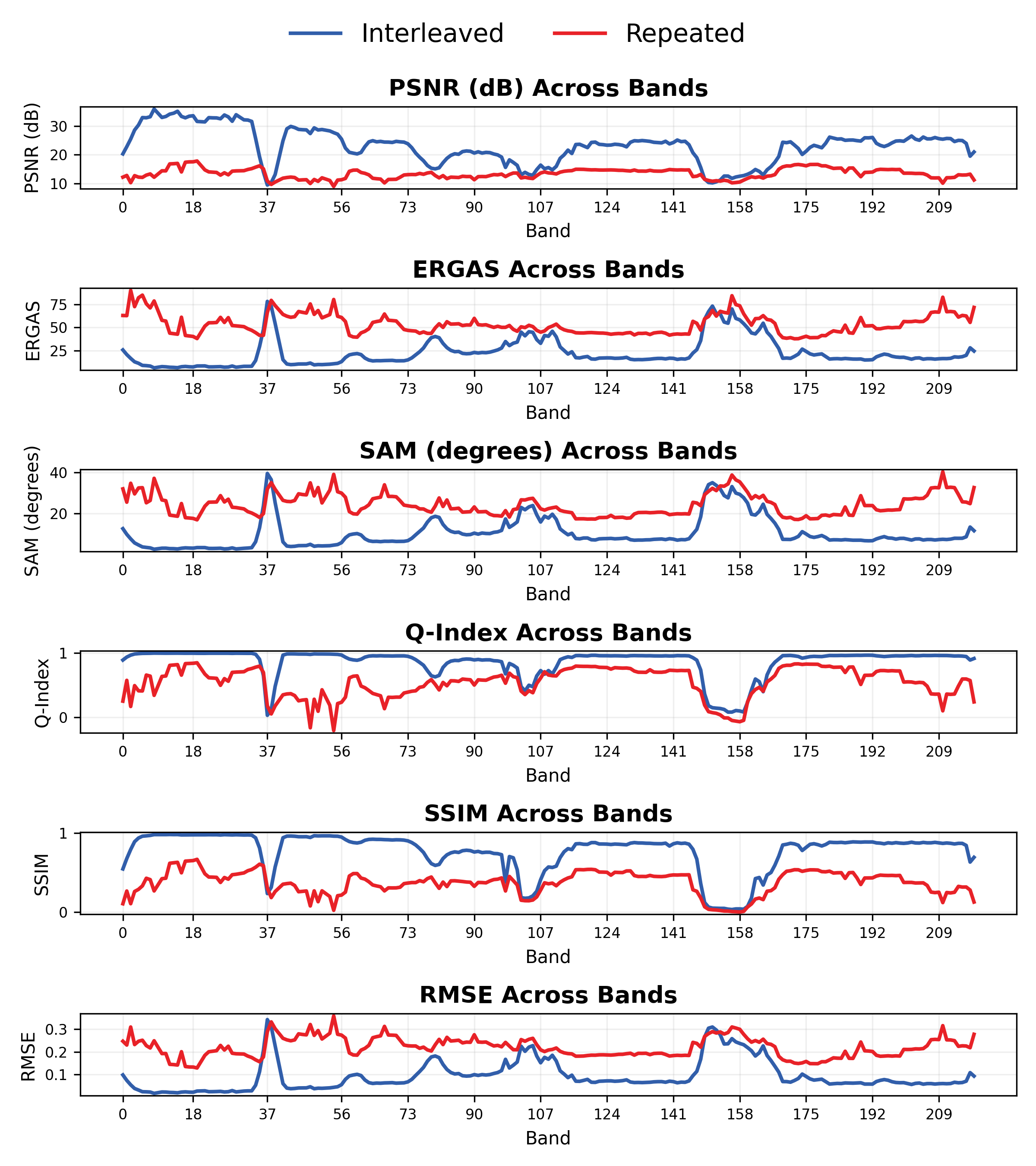}
    \caption{Quantitative comparison of two band sequence designs—\textbf{Interleaved} ([30, 20, 9, 40, 52, 20, 40, 30, 9, 52, 30, 20, 9, 40, 52, 30]) and \textbf{Repeated} ([30, 30, 30, 30, 20, 20, 20, 9, 9, 9, 40, 40, 40, 52, 52, 52])—evaluated across all spectral bands using six metrics: PSNR, ERGAS, SAM, Q-Index, SSIM, and RMSE. Each subplot shows the average metric values across 30 test samples. Higher values of PSNR, Q-Index, and SSIM, and lower values of ERGAS, SAM, and RMSE indicate better reconstruction performance.}
    \label{fig:physical_repeated}
\end{figure}

As illustrated in \autoref{fig:architecture}, the input depth requires multiples of 32, while the shifted window size is 7×7×7. To accommodate this requirement, we constructed a 16-depth input and concatenated it with itself to achieve a 32-depth input. Utilizing transformer positional embedding \cite{liu2021swin}, the Swin Transformer block effectively processes 7 depths of the input. We conducted a comparative analysis of two distinct band sequence designs. The first design employs an interleaved approach [30, 20, 9, 40, 52, 20, 40, 30, 9, 52, 30, 20, 9, 40, 52, 30], which ensures that all unique band pairs co-occur within at least one adjacent slice pair in the depth dimension. The second design implements a contrast repeated pattern [30, 30, 30, 30, 20, 20, 20, 9, 9, 9, 40, 40, 40, 52, 52, 52], maintaining equal frequency of each band while presenting them in a repeated sequential order. Performance metrics across six evaluation criteria are presented in \autoref{fig:physical_repeated}. The results demonstrate that the interleaved design consistently outperforms the repeated design across nearly all bands. This superior performance suggests that the Swin Transformer mechanism significantly enhances model efficacy by effectively capturing and appropriately weighting attention for each 7×7 patch in the depth dimension, thereby optimizing the representation of band and spectral information.

\subsection{Different Cascade Training Strategies}
\begin{figure}[H]
    \centering
    \includegraphics[width=1\columnwidth]{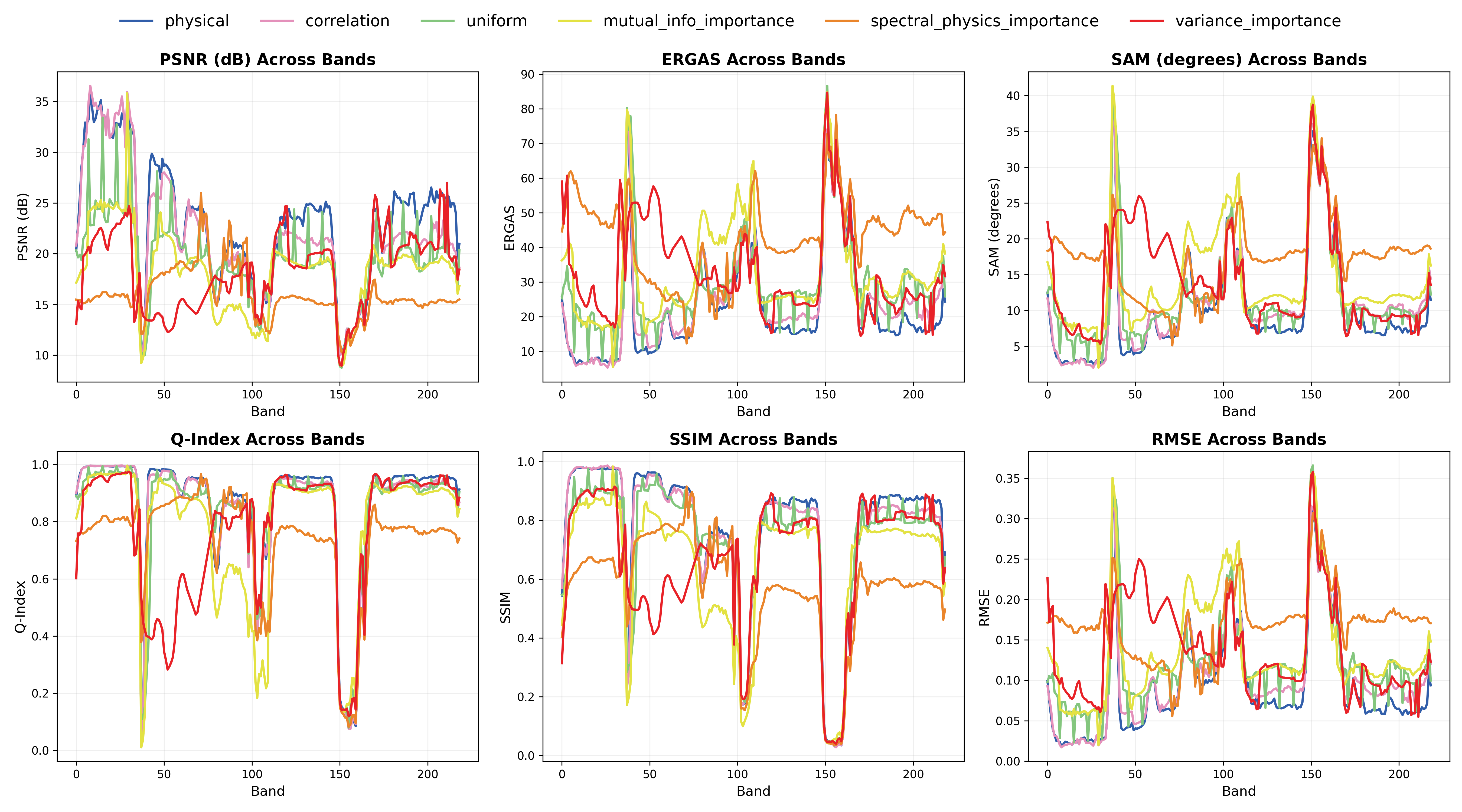}
    \caption{Quantitative comparison of six cascade training strategies—physical, correlation\_importance, uniform, mutual\_info\_importance, spectral\_physics\_importance, and variance\_importance—evaluated across all spectral bands using six metrics: PSNR, ERGAS, SAM, Q-Index, SSIM, and RMSE. Each subplot shows the average metric values across 30 test samples. Higher values of PSNR, Q-Index, and SSIM, and lower values of ERGAS, SAM, and RMSE indicate better reconstruction performance.}
    \label{fig:6basics}
\end{figure}

To balance performance and computational resources, we employed a two-stage fine-tuning approach. In the first stage, we constructed cascade training pyramids comprising 29 bands, followed by fine-tuning the remaining hyperspectral bands in the second stage. \autoref{tab:cascade_strategy} presents the cascade training pyramids for different bands, while \autoref{fig:6basics} illustrates the performance metrics of six distinct cascade training strategies. As discussed in \autoref{fig:band_region_overall} and \autoref{sec:result}, Area II (distant wavelength area) exhibited diminished performance due to its limited correlation with multispectral band generation. Therefore, designing an effective cascade training strategy is crucial.
We evaluated six different cascade training strategies:
\textbf{Uniform strategy}: No prior knowledge implementation. Progressive construction from Area I (multispectral generating bands) to Area II (hyperspectral generated bands).
\textbf{Physics strategy}: Incorporates physics-based prior knowledge. Constructs cascade pyramid based on physical spectral ranges and remote sensing bands (RGB, NIR, SWIR1, SWIR2).
\textbf{Variance-based importance}: Data-driven approach quantified by:
$\text{Var}(b) = \frac{1}{N} \sum_{i=1}^N (x_{i,b} - \mu_b)^2$
where $x_{i,b}$ represents the pixel value of band $b$ in the $i$-th sample, and $\mu_b$ is the mean value of band $b$. Bands with higher variance are prioritized for reconstruction.
\textbf{Correlation-based importance}: Data-driven approach using Pearson correlation coefficient:
$r_{b,s} = \frac{\text{Cov}(X_b, X_s)}{\sigma_b \sigma_s}$
where $X_b$ represents target band pixels, $X_s$ represents input band pixels, $\text{Cov}$ is covariance, and $\sigma$ is standard deviation. For each target band $b$, we calculate:
$\text{CorrImp}(b) = \max_{s \in S} |r_{b,s}|$
Bands are then prioritized by correlation magnitude.
\textbf{Mutual information importance}: Data-driven approach using mutual information:
$I(b, s) = \sum_{x_b, x_s} p(x_b, x_s) \log \frac{p(x_b, x_s)}{p(x_b)p(x_s)}$
For each target band $b$, we calculate:
$\text{MIImp}(b) = \max_{s \in S} I(b, s)$
Bands with higher mutual information are prioritized.
\textbf{Spectral-physics importance}: Combined data-physics approach that integrates physical band segmentation (visible light, near-infrared, short-wave infrared) with inter-band distances, prioritizing bands with stronger physical significance.
\par
Intuitively, one might expect that focusing on Area II bands by positioning them in the earlier levels of the cascade pyramid (levels 0 or 1), as in the variance-based strategy, or distributing them uniformly across levels, would yield superior results. However, our experimental results indicate that the physics-based and correlation-based strategies consistently outperformed other approaches across all bands. This suggests that our designed strategies successfully capture the intrinsic spectral information and inter-band correlations—not only between multispectral and hyperspectral bands but also among hyperspectral bands at various cascade training levels.

\section{Applications to Downstream Tasks}
The SpecSwin3D model demonstrates significant versatility in downstream applications, addressing limitations in both spatial and spectral resolution domains. In terms of spatial resolution enhancement, the model facilitates the generation of high-resolution hyperspectral bands by leveraging the inherently superior spatial resolution of multispectral generating bands. This capability substantially improves hyperspectral-based applications through enhanced spatial detail. Concurrently, in the spectral resolution domain, the model augments multispectral-based applications by generating intermediate hyperspectral bands, thereby increasing spectral resolution and subsequently enhancing application performance through more comprehensive spectral information capture.
\par
We use the Sentinel-2 dataset for downstream tasks, which includes 10 m resolution bands: Band 2 (Blue, 490 nm center, 65 nm bandwidth), Band 3 (Green, 560 nm, 35 nm), Band 4 (Red, 665 nm, 30 nm), and Band 8 (NIR, 842 nm, 115 nm); and 20 m resolution bands: Band 5 (Red Edge 1, 705 nm, 15 nm), Band 6 (Red Edge 2, 740 nm, 15 nm), Band 7 (Red Edge 3, 783 nm, 20 nm), Band 8A (Narrow NIR, 865 nm, 20 nm), Band 11 (SWIR 1, 1610 nm, 90 nm), and Band 12 (SWIR 2, 2190 nm, 180 nm). 
\par
The \textbf{SpecSwin3D} models were trained using Sentinel-2 10 m bands (2, 3, 4, 8, with band 8 duplicated) as a 3D input sequence.

\subsection{Hyperspectral-based Application: Landuse Segmentation}
We used the public DynamicWorld dataset \cite{brown2022dynamic} as our ground truth, which provides landuse labels with 9 classes. We reclassified these into 4 categories: water, vegetation, farmland, and built environment (buildings, bare land, and roads). For classification, we employed the CatBoost algorithm  \cite{prokhorenkova2018catboost}. We first performed land use classification using a combined Sentinel-2 feature set that includes the 20 m bands (B5, B6, B7, B11, B12) together with the 10 m bands (B2, B3, B4, B8). To ensure a common spatial grid, the 10 m bands were downsampled to 20 m. To enhance spatial detail, we then applied \textbf{SpecSwin3D}, which takes the native 10 m bands (B2, B3, B4, B8; with B8 duplicated) as input and generates 10 m hyperspectral bands matching the central wavelengths and bandwidths of the original 20 m bands (B5--B7, B11--B12).
\par
\begin{figure}[htbp!]
\begin{center}
		\includegraphics[width=0.8\columnwidth]{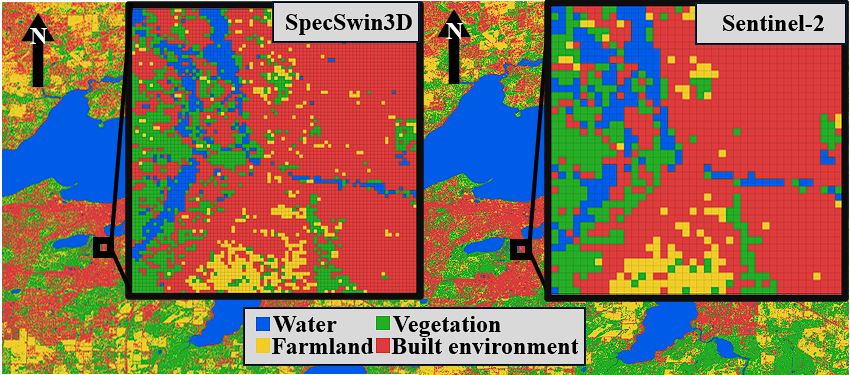}
	\caption{Land-use segmentation results. Bottom: using Sentinel-2 imagery (20\,m bands). Top: using SpecSwin3D (10\,m bands). Both panels depict the same area in Madison, Wisconsin; colors denote Water (blue), Vegetation (green), Farmland (yellow), and Built environment (red). The generated hyperspectral bands maintain comparable classification performance while substantially enhancing spatial resolution, yielding clearer boundary delineation.}
\label{fig:landuse}
\end{center}
\end{figure}

The results (\autoref{fig:landuse}), shown in \autoref{tab:ds_result}, demonstrate
that the generated hyperspectral bands achieved comparable classification
performance to the 20\,m Sentinel-2 baseline (72.1\% vs.\ 74.3\%), while simultaneously
providing a substantial improvement in spatial resolution from 20\,m to 10\,m.
This finer granularity is especially valuable for delineating boundaries between
farmland and built-up areas, which are often poorly resolved at coarser resolutions.
\par
Although the overall accuracy of land use classification with SpecSwin3D-generated
bands (72.1\%) is slightly lower than that obtained using Sentinel-2 20\,m bands
(74.3\%), the difference remains small and within a satisfactory range.
More importantly, SpecSwin3D provides hyperspectral representations at 10\,m
resolution, which deliver much finer spatial detail compared with the 20\,m baseline.
This improvement in spatial granularity is particularly valuable for distinguishing
boundaries between farmland and built-up areas, and it enables subsequent downstream
tasks to benefit from the enhanced detail despite a marginal drop in overall accuracy.
\begin{figure}[htbp!]
\begin{center}
		\includegraphics[width=0.8\columnwidth]{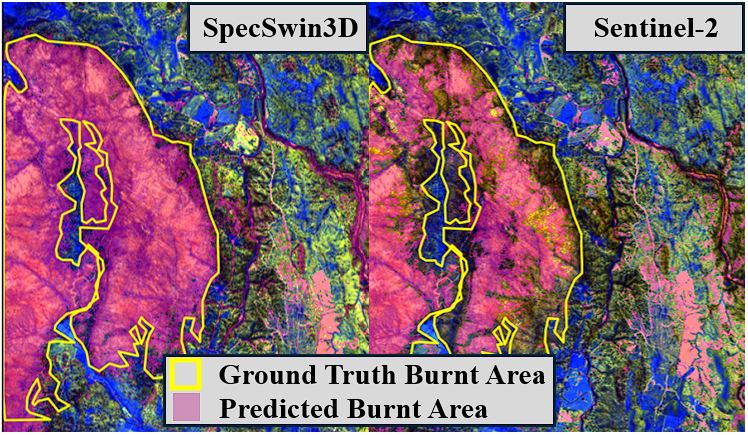}
	\caption{Burnt-area detection results. Left: SpecSwin3D (dNBR). Right: Sentinel-2 (dNDVI). Yellow outlines indicate the ground-truth burnt perimeter; magenta shading shows the predicted burnt area. Both panels depict the same location. Spectral enrichment via SpecSwin3D-generated SWIR2 enables NBR computation, dramatically improving burnt area detection accuracy and recall over NDVI-only approaches.}

\label{fig:burntarea}
\end{center}
\end{figure}
\subsection{Multispectral-based Application: Burnt Area Detection}
We detect burnt areas using the BiAU-Net dataset \cite{sui2024biau}, a bi-temporal burnt area dataset based on Sentinel-2 imagery with 10 m bands (Bands 2, 3, 4, and 8) captured before and after wildfire events. For detection, we apply a threshold-based method using the difference in NDVI between pre- and post-fire images (NDVI\textsubscript{pre} -- NDVI\textsubscript{post} > threshold). To further enhance detection, we employ the \textbf{SpecSwin3D} model to generate hyperspectral bands in the SWIR2 region, enabling accurate computation of the NBR difference (NBR\textsubscript{pre} -- NBR\textsubscript{post} > threshold) \cite{escuin2008fire,wu2024remote}. The results (\autoref{fig:burntarea}) are shown in \autoref{tab:ds_result}, demonstrating that the inclusion of generated hyperspectral bands significantly improves the effectiveness of the threshold-based approach.
\par
Although NDVI-based methods using only Sentinel-2 bands suffer from limited
sensitivity to burnt areas (83.1\% accuracy with recall of 55.7\%), the addition of
SpecSwin3D-generated SWIR2 bands enable computation of NBR, boosting performance
to 94.1\% accuracy with recall of 92.7\%. This substantial gain highlights the
importance of spectral enrichment: by providing access to wavelengths not available
in the original Sentinel-2 inputs, SpecSwin3D allows even simple threshold-based
approaches to achieve state-of-the-art burnt area detection.

\begin{table}[htbp]
\centering
\caption{Performance comparison in applications}
\label{tab:ds_result}
\small
\setlength{\tabcolsep}{3pt}
\begin{tabular}{|l|c|c|c|c|c|}
\hline
\textbf{Dataset} & \makecell{\textbf{Spatial} \\ \textbf{Resolution}}
 & \textbf{Accuracy} & \textbf{Precision} & \textbf{Recall} & \textbf{F1 Score} \\
\hline
\multicolumn{6}{|c|}{\makecell{\textit{Hyperspectral-based: Landuse Segmentation (Average of classes)}}} \\
\hline
Sentinel-2 & 20 m & 74.3\% & 77.9\% & 78.7\% & 78.2\% \\
SpecSwin3D & 10 m & 72.1\% & 72.3\% & 72.1\% & 72.1\% \\
\hline
\multicolumn{6}{|c|}{\makecell{\textit{Multispectral-based: Burnt Area Detection}}} \\
\hline
Sentinel-2 & 10 m & 83.1\% & 90.1\% & 55.7\% & 68.8\% \\
SpecSwin3D & 10 m & 94.1\% & 90.0\% & 92.7\% & 91.3\% \\
\hline
\end{tabular}
\end{table}

\section{Conclusion and Future Work}
Our proposed model, \textbf{SpecSwin3D}, effectively captures the underlying physical and spectral characteristics of reflectance across hyperspectral bands and reveals the latent relationships between multispectral and hyperspectral data. By integrating a U-Net architecture equipped with the 3D transformer-based encoder, SpecSwin3D jointly learns spatial and spectral representations, leading to performance gains over existing state-of-the-art methods.
\par
\subsection{Band Sequence Optimization and Input Depth Considerations}
As discussed in \autoref{sec:sequence}, our model architecture requires input depths that are multiples of 16 (or 32 when bypassing the self-concatenation step). To ensure comprehensive spectral information capture, all unique band pairs must co-occur within at least one adjacent slice pair. The minimum band sequence length to satisfy this condition is 11, leaving 5 available depth positions. In our implementation, we utilized the bands themselves as padding for these remaining positions.
\par
We conducted experiments comparing this approach against utilizing established remote sensing indices—Normalized Difference Vegetation Index (NDVI), Normalized Difference Water Index (NDWI), Normalized Difference Built-up Index (NDBI), Normalized Difference Snow Index (NDSI), and Transformed Vegetation Index (TVI)—as padding elements. Results demonstrated that the Swin Transformer mechanism, which employs complex depth processing beyond simple concatenation and 3D window shifting, exhibited significantly reduced performance when utilizing indices compared to band self-padding.
\par
Future research directions include investigating longer and more sophisticated band sequence designs that incorporate both the five generating multispectral bands and various remote sensing indices. Additionally, while the current 3D shifted window size of 7×7×7 enables the model to consider seven depth slices simultaneously, our current analysis focuses solely on single-depth distances between band pairs. Subsequent studies will extend this to incorporate multi-depth distance relationships, potentially enhancing spectral information integration.
\subsection{Cascade Strategy Design}
In this study, we implemented different cascade strategies through deterministic, hard-coded methodologies. This approach demonstrated significant computational efficiency, with approximately 80\% of bands requiring only 40\% of the training epochs through our fine-tuning protocol. These strategies were primarily derived from spectral and physical priori experience or data-driven statistical analyses of pixel values across spectral bands. However, our empirical evaluations revealed that such predetermined strategies do not consistently yield optimal performance across diverse environmental contexts. Comparative analyses between urban and agricultural landscapes demonstrated that models exhibiting superior performance in one domain occasionally produced suboptimal results for specific spectral bands in alternative domains. This observation suggests the necessity for developing adaptive learning frameworks capable of autonomously determining optimal cascade strategies based on the inherent spectral characteristics of different land cover types. Future research will focus on developing a self-optimizing pipeline that enables the model to learn and adapt cascade strategies according to specific spectral and spatial contexts.
\subsection{Depth vs Channel}
Our approach employs depth dimensionality to represent distinct wavelength bands in remote sensing spectrum, utilizing 3D shifting windows to effectively capture both inter-band and intra-band relationships. While spectral information can be encoded in either depth or channel dimensions, these approaches create a critical trade-off. In the depth-oriented setting, positional embedding information is preserved through the 3D Swin Transformer architecture, which helps SpecSwin3D perform better. Meanwhile, the channel-oriented setting—though computationally more efficient (faster training and smaller model)—sacrifices performance because it operates as a 2D input.
\par
This presents a critical trade-off between computational efficiency and model performance. To address this challenge, our SpecSwin3D framework incorporates a dimensional parameter mechanism (3D or 2D) that allows users to make case-by-case decisions

\section*{Code and data availability}
All code and datasets used in this study are available at
\url{https://github.com/scdmlab/SpecSwin3D}.

\bibliographystyle{elsarticle-num} 
\bibliography{refs}

\end{document}